\documentclass[sigconf]{acmart} 

\setcopyright{acmcopyright} 
\acmYear{2025} 
\copyrightyear{2025} 
\acmConference[ICAIL 2025]{International Conference on Artificial Intelligence and Law 2025}{June 16 - 20, 2025}{Chicago, IL} 
\usepackage{appendix}
\usepackage{framed,graphicx,xcolor}
\colorlet{shadecolor}{orange!15}
\usepackage{mdframed}
\usepackage{subcaption}
\begin{document} 
\title[Identifying Legal Holdings with LLMs]{Identifying Legal Holdings with LLMs: A Systematic Study of Performance, Scale, and Memorization} 

\begin{abstract}
As large language models (LLMs) continue to advance in capabilities, it is essential to assess how they perform on established benchmarks. In this study, we present a suite of experiments to assess the performance of modern LLMs (ranging from 3B to 90B+ parameters) on \texttt{CaseHOLD}, a legal benchmark dataset for identifying case holdings. Our experiments demonstrate ``scaling effects'' - performance on this task improves with model size, with more capable models like \texttt{GPT4o} and \texttt{AmazonNovaPro} achieving macro F1 scores of 0.744 and 0.720 respectively. These scores are competitive with the best published results on this dataset, and do not require any technically sophisticated model training, fine-tuning or few-shot prompting. To ensure that these strong results are not due to memorization of judicial opinions contained in the training data, we develop and utilize a novel citation anonymization test that preserves semantic meaning while ensuring case names and citations are fictitious. Models maintain strong performance under these conditions (macro F1 of 0.728), suggesting the performance is not due to rote memorization. These findings demonstrate both the promise and current limitations of LLMs for legal tasks with important implications for the development and measurement of automated legal analytics and legal benchmarks.
\end{abstract}

\author{Chuck Arvin}
\authornote{This work does not relate to the author's position at Amazon. All views expressed are the author's own.}
\authornote{Generative AI disclosure: Claude was used to provide copy-editing feedback and to improve the code used for data visualization.}
\email{carvin@usc.edu}
\affiliation{%
  \institution{USC Gould School of Law}
  \city{Los Angeles}
  \state{California}
  \country{USA}
}


\maketitle

\section{Introduction}

Large language models have demonstrated remarkable capabilities across diverse domains, from machine translation \cite{zhu2024multilingualmachinetranslationlarge, feng2024tearimprovingllmbasedmachine}, code development \cite{wang2024planningnaturallanguageimproves, koziolek2024llm}, and writing assistance \cite{liang2024mappingincreasingusellms, stahl2024exploring}. Their application to legal analysis is especially promising, as the legal domain is a heavily text-driven one. Legal documents - including contracts, briefs, and judicial decisions - rely on specialized language, including ``terms of art'' and ``extreme precision of expression'' \cite{mellinkoff2004language, katz2023naturallanguageprocessinglegal}. Further, the effective practice of law requires the ability to process and understand rich sources of unstructured textual information. Improvements in this space may transform key aspects of legal practice: from accelerating e-discovery \cite{mit-ediscovery} and enhancing document drafting \cite{villasenor2024} to automating case summarization \cite{shen2022multilexsumrealworldsummariescivil}.

However, legal applications for LLMs come with high ethical and professional standards. For example, state bar associations have released clarifications highlighting the professional requirements for attorneys using generative AI technologies \cite{calibar}. Undetected issues or hallucinations may hinder the administration of justice, as individuals may receive incompetent representation, responsive documents may not be turned over, or answers may reflect inherent bias. Thus, despite the potential upside, successful application of LLMs to the legal domain requires thoughtful measurement to ensure the models are producing accurate and honest answers.

In order to improve our ability to measure the quality of these answers, we have seen a proliferation of legal benchmarks. These benchmarks allow practitioners to rigorously evaluate if their models produce correct answers and to root out failure modes. For example, the \texttt{CaseHOLD} dataset allows researchers to systematically assess how well their models can identify the key legal holding in a case \cite{zheng2021doespretraininghelpassessing}. The \texttt{LegalBench} dataset contains numerous hand-labeled and automated datasets for various legal tasks, including assessment of which laws apply to a particular fact pattern \cite{guha2024legalbench}.

However, there are subtle risks to these kinds of benchmarks. The \texttt{CaseHOLD} dataset, the broader \texttt{LexGLUE} dataset, and others may be built on publicly available judicial opinions \cite{chalkidis2022lexgluebenchmarkdatasetlegal}. These decisions provide a readily available source of high-quality judicial reasoning, and do not require expensive human-labeling efforts. But modern LLMs are likely trained on the same corpus of text. Recent research has shown that LLMs are capable of ``memorizing'' their training data \cite{hartmann2023sokmemorizationgeneralpurposelarge, henderson2022pilelawlearningresponsible}. In a notable public example, the New York Times demonstrated that ChatGPT was capable of reproducing more than 100 published articles \cite{newyorktimes2024}. This presents a critical question for researchers: when an LLM performs well on a given task, is it simply reciting memorized text it saw during training? 

In this paper, we analyze how modern LLMs perform in identifying case holdings using the \texttt{CaseHOLD} dataset \cite{zheng2021doespretraininghelpassessing}. The dataset consists of 5,314 observations, measuring a key component of legal reasoning - the ability to summarize a legal decision into a concise and relevant legal holding. The holding offers a summary of the legal rule established or applied in a case. We evaluate LLMs of varying model sizes, ranging from 3 billion to 90+ billion parameters, and utilize a novel citation anonymization technique to detect if the LLMs are simply engaging in rote memorization. 

In addition to our code and datasets\footnote{\href{https://github.com/chuck-arvin/CaseHOLD2025}{https://github.com/chuck-arvin/CaseHOLD2025}}, our work contributes three key findings to the literature:
\begin{itemize}
    \item Modern LLMs can perform competitively with custom-built legal models on the \texttt{CaseHOLD} benchmark without fine-tuning or domain adaptation. \texttt{GPT4o} achieves a macro F1 score of 0.742 - this score outperforms three of the five custom legal models trained in \cite{niklaus2024_legalpile}, which reports macro F1 scores ranging from 0.717 to 0.770.
    \item Performance on this task scales with model size, across the Llama, Amazon Nova, and GPT4o model families. This ``scaling effect'' mimics those seen elsewhere in the literature, and suggest that as LLMs continue to improve on general purpose tasks, these models may achieve even stronger results on legal tasks like this one.
    \item We propose a novel citation anonymization technique which may be applied to other legal NLP tasks. Strong performance (macro F1 of 0.728) remains even after we introduce ``anonymized citations'', suggesting that the LLMs are not engaging in rote memorization of their training data. 
\end{itemize}

In Section \ref{sec:design}, we discuss our dataset, research design and empirical results demonstrating that LLMs perform well on this task in a zero-shot manner. In Section \ref{sec:memorization}, we introduce our citation anonymization methodology, and show that our results are robust to large changes to the inputs. Finally, we conclude with a discussion of our results and future research directions.

\section{Research Design}
\label{sec:design}
\subsection{\texttt{CaseHOLD} Dataset}
Our work utilizes the \texttt{CaseHOLD} dataset, first published in \cite{zheng2021doespretraininghelpassessing} and incorporated in \texttt{LexGLUE} \cite{chalkidis2022lexgluebenchmarkdatasetlegal}. This dataset aims to simulate the task of ``identifying the legal holding of a case'', turning this task into a multiple choice Q\&A dataset with clear success metrics. In future work, we plan to expand this analysis to broader legal NLP datasets.

Figures \ref{fig:exampleone} and \ref{fig:exampletwo} present two questions from this dataset of varying difficulty. The input prompt is on the left, and the model must identify the correct choice to complete the parenthetical \textbf{<HOLDING>} citation. On the right are the five answer choices, with the correct answer choice highlighted in \textcolor{blue}{blue}, and the models which selected that answer in brackets. In Figure \ref{fig:exampleone}, key terms like ``Coast Guard'' and ``vessel'' hint at the correct answer, while in Figure \ref{fig:exampletwo}, the model must select from multiple legal doctrines, including jurisdictional questions and multiple flavors of the ``filed rate'' doctrine.

\begin{figure}
\begin{mdframed}
\begin{minipage}{0.45\textwidth}
\footnotesize and suppressing violations of laws of the United States, “officers may at any time go on board of any vessel subject to the jurisdiction, or to the operation of any law, of the United States, address inquiries to those on board, examine the ship’s documents and papers, and examine, inspect, and search the vessel and use all necessary force to compel compliance.” This statute has been construed to permit the Coast Guard to stop an American vessel in order to conduct “a document and safety inspection on the high seas, even in the absence of a warrant or suspicion of wrongdoing,” United States v. Hilton, 619 F.2d 127, 131 (1st Cir.1980), and to conduct a more intrusive search on the basis of reasonable suspicion, see United States v. Wright-Barker, 784 F.2d 161, 176 (3d Cir.1986) (\textbf{<HOLDING>}), superseded by statute on other grounds as
\end{minipage}\hfill
\begin{minipage}{0.45\textwidth}
\footnotesize 
\begin{itemize}
\item holding that forfeiture statute is subject to the fourth amendments prohibitions against unreasonable searches and seizures
\item holding that the fourth amendment proscription against unreasonable searches and seizures was applicable to the states under the fourteenth amendment so that evidence seized in violation of the constitution could no longer be used in state courts
\item holding sbm is not a violation of the defendants fourth amendment right to be free from unreasonable searches and seizures
\item \textcolor{blue}{holding that a reasonable suspicion requirement for searches and seizures on the high seas survives fourth amendment scrutiny}[\textbf{ALL MODELS}]
\item holding that inmates fourth amendment protection from unreasonable strip searches survives hudson
\end{itemize}
\end{minipage}
\caption{All models agree on the correct answer. As GPT4o states, ``it directly addresses the Fourth Amendment scrutiny of searches and seizures on the high seas''.}
\label{fig:exampleone}
\end{mdframed}
\end{figure}

\begin{figure}
\begin{mdframed}
\begin{minipage}{0.45\textwidth}
\footnotesize If the trial discloses that the services were not formally ordered, then the discussion herein concerning constructive ordering of services is applicable. 12 . See United Artists Payphone Corp. v. New York Tel. Co., 8 FCC Rcd 5563 (1993); see also AT \& T v. City of New York, 83 F.3d 549 (2d Cir.1996); AT \& T Corp. v. Community Health Group, 931 F.Supp. 719 (S.D.Cal.1995); In re Access Charge Reform, 14 FCC Rcd 14221 (1999). 13 . See also City of New York, 83 F.3d at 553 (relying on United Artists to hold that a party can become a " ‘customer,’ in one of two ways: (1) by affirmatively ordering the service ... or (2) by constructively ordering [service] and creating an 'inadvertent carrier-customer relationship ....' ”). 14 . See US Wats, Inc. v. AT \& T Co., 1994 WL 116009, at *5 (\textbf{<HOLDING>}). 15 . While not directly stating so, AT \& T
\end{minipage}\hfill
\begin{minipage}{0.45\textwidth}
\footnotesize
\begin{itemize}
\item \textcolor{blue}{holding that filedrate doctrine does not apply when adjudication of the plaintiffs claim will not result in rate discrimination nor embroil the court in a dispute over reasonableness of charges} [\textbf{\texttt{NovaLite}, \texttt{NovaPro}, \texttt{GPT4o-mini}, \texttt{GPT4o}}]

\item holding that the doctrine does not apply in such circumstancesi [\textbf{\texttt{Llama3.2-3B}}]

\item holding that this court does not have jurisdiction over plaintiffs claims because the court may review neither criminal matters nor the decisions of district courts

\item holding that under the filed rate doctrine a question regarding reasonable rates should be addressed to the department of insurance and that the rate plaintiff was charged is conclusively presumed reasonable under the filed rate doctrine [\textbf{\texttt{Llama3.2-11B}, \texttt{Llama3.2-90B}}]

\item holding the same for the other separate rate plaintiffs in this action
\end{itemize}
\end{minipage}
\caption{The models reach three distinct answers. As \texttt{NovaPro} states, ``Given the context and the need for a holding that aligns with the discussion of service ordering and customer relationships, Option A is the most appropriate as it addresses a relevant doctrine in a manner consistent with the input text''}
\label{fig:exampletwo}
\end{mdframed}
\end{figure}

\subsection{Experimental Design}
In our first experiment, we test the zero-shot abilities of modern large language models. We use a standardized Prompt \ref{fig:prompt1} to ask the LLM to analyze the surrounding text, evaluate the five multiple-choice options, and finally decide which one best completes the passage. We ask the LLM to ``reason'' about the right replacement text before answering. This is an example of \textit{Chain-of-Thought} prompting, which has been shown to improve LLM performance on reasoning tasks \cite{wei2023chainofthoughtpromptingelicitsreasoning}. This also mirrors the way a human might approach these questions - thinking through the question and examining the answer choices before producing a final answer choice. 

\begin{figure}
\begin{shaded}
\raggedright
\footnotesize Task: Legal Holding Identification
    
    Context: You are analyzing a legal text to identify the most appropriate legal holding. A legal holding is the court's determination of a matter of law based on the facts of a particular case.
    
    Input Text: \texttt{citing prompt}

    Question: Based on the legal context above, which of the following holdings best completes the text where the <HOLDING> tag appears? Consider:\\
    - The specific legal issue being discussed\\
    - The logical flow of the legal argument \\
    - The precedential value implied by the context\\

    Options:\\
    A: \texttt{holding 0}\\
    B: \texttt{holding 1}\\
    C: \texttt{holding 2}\\
    D: \texttt{holding 3}\\
    E: \texttt{holding 4}\\

    Instructions:\\
    1. Analyze the context and legal reasoning in the input text\\
    2. Consider how each option would fit within the legal argument\\
    3. Evaluate which option best maintains the logical flow. Explain your reasoning first, formatted like this <reasoning> reasoning </reasoning>\\
    4. Provide your final answer in the format: ANSWER: X (where X is A, B, C, D, or E)
\end{shaded}
\captionsetup{name=Prompt}
\caption{This prompt asks an LLM to read the citing text, analyze the options, and conclude with the best fitting option.}
\label{fig:prompt1}
\end{figure}

We run this prompt on a suite of 8 LLMs of varying sizes and capabilities (\texttt{AmazonNovaMicro}, \texttt{AmazonNovaLite}, \texttt{AmazonNovaPro}, \texttt{Llama3.2-3B}, \texttt{Llama3.2-11B}, \texttt{Llama3.2-90B}, \texttt{GPT4o-mini}, \texttt{GPT4o}) \cite{amazonnova, llama, openai2024gpt4ocard}. These models represent several families of modern LLMs. All experiments are run in a zero-shot manner - we do not provide examples or fine-tune the models on relevant data. To ensure comparability with published results, we label the entirety of the \texttt{CaseHOLD} test dataset, 5,314 observations. LLMs are run at temperature 0 to obtain deterministic results. We parse the LLM responses\footnote{Due to inconsistencies in how each LLM formats its response, we use the following regex to extract an answer. \texttt{{\textbackslash}bANSWER:{\textbackslash}s*([A-E])}. In words, we identify the word ANSWER:, then extract the following single character [A - E].}. In the rare (< 1\%) of situations where a response is not parseable, we randomly select an answer choice from the set. 

\subsection{Zero-Shot Empirical Results}
\label{sec:results}

\begin{figure}
    \centering
    \includegraphics[width=0.8\linewidth]{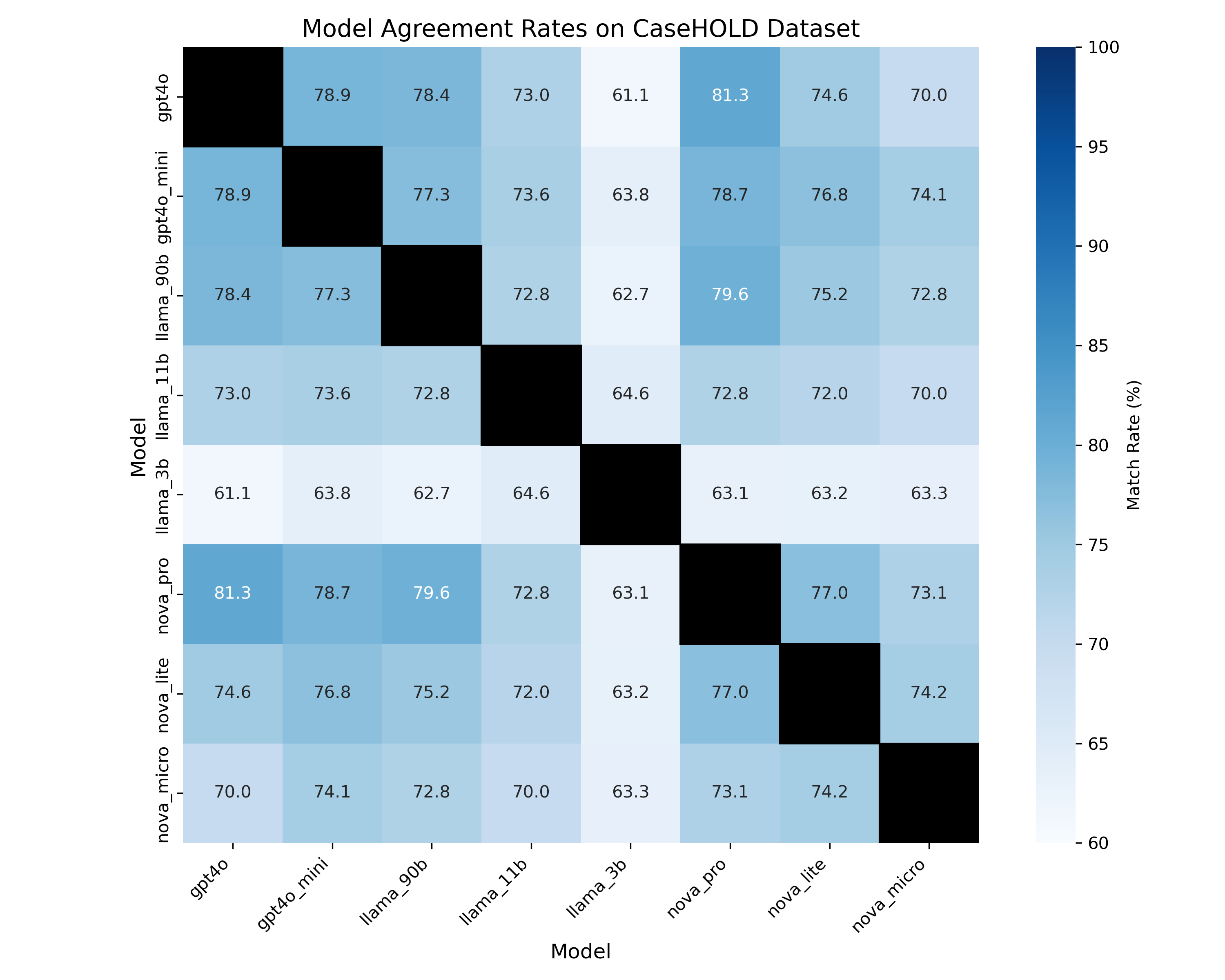}
\caption{Match rate for answers between different models. LLMs frequently agree on the correct answer.}
    \label{fig:matchrate}
\end{figure}

First, we measure the diversity of the model responses. When two models respond with the same answer to a given question, we code that as a match. The match rate measures how often two models give the same answer. In Figure \ref{fig:matchrate}, we show the match rate between the models. For example, the \texttt{GPT-4o} and \texttt{NovaMicro} models produced the same answer in 70.0\% of cases. All model pairs produced the same answer more than 50\% of the time. 

We measure the number of unique choices to a given question. Table \ref{tab: matchrate} shows the number of unique choices for each question. In 34\% of cases, all models tested chose the same answer - here, the models identified the correct choice more than 90\% of the time. In other cases, the models split on the correct answer. Accuracy for questions with two or more different answers was much lower.

\begin{table}[h!]
\begin{tabular}{r|rr}
\toprule
\# of Unique Choices & Accuracy & \# of Questions \\
\midrule
1 & 92.1\% & 2165 \\
2 & 58.0\% & 1987 \\
3 & 40.4\% & 923 \\
4 & 29.9\% & 221 \\
5 & 17.4\% & 18 \\
\bottomrule
\end{tabular} 
\caption{In many cases, all LLMs tested gave the same answer. These questions showed very high accuracy. Questions with multiple unique responses showed lower accuracy.}
\label{tab: matchrate}
\end{table}

We compute the macro F1 score which measures both precision (the fraction of correct predictions among all predictions of a given class) and recall (the fraction of actual instances of a class that were correctly identified). The macro F1 score averages the F1 score across all five answer choices to account for imbalance in how often each appears. Higher values are better, with 1 representing perfect performance. We include the best performing reported results from three papers examining the \texttt{CaseHOLD} dataset \cite{zheng2021doespretraininghelpassessing, chalkidis2022lexgluebenchmarkdatasetlegal, niklaus2024_legalpile}. 
\begin{figure}
\centering
\includegraphics[width=0.9\linewidth]{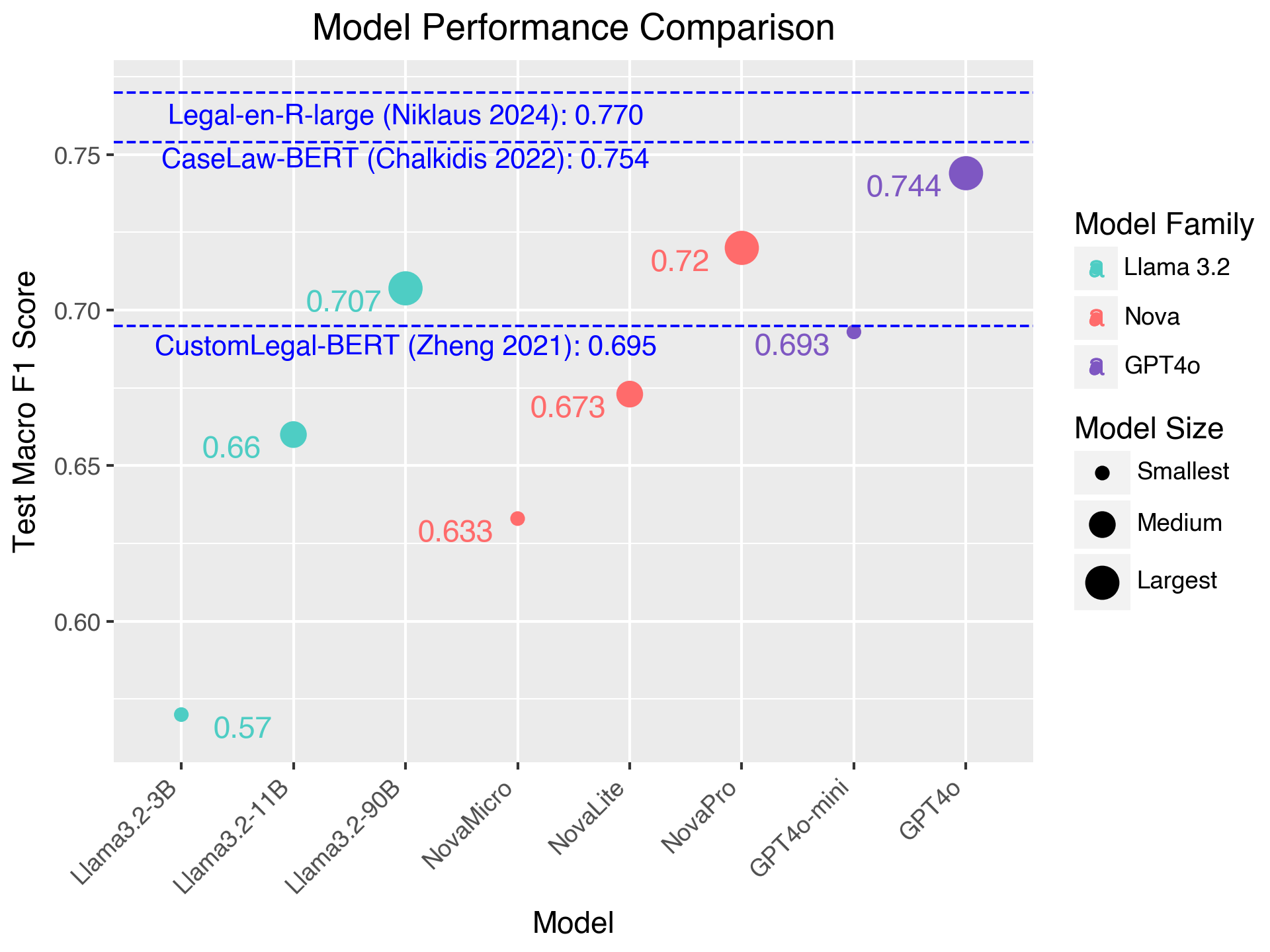}
\caption{Macro F1 Scores on the CaseHOLD test set. Model performance improves with model size across all model families tested.}
    \label{fig:f1-scores}
\end{figure}

A few salient observations given these results:
\begin{itemize}
    \item As Figure \ref{fig:f1-scores} shows, performance scales with model size. Among the \texttt{Llama}, \texttt{Nova} and \texttt{GPT-4o} model families, performance on this task improves with parameter size. This aligns with the scaling laws seen elsewhere in the literature (\cite{brown2020languagemodelsfewshotlearners, kaplan2020scalinglawsneurallanguage}), and suggests that improvements to general purpose LLMs may improve performance on specific legal tasks.

    \item Several models match or surpass the performance of custom-built legal models. \texttt{GPT-4o} performance of 0.744 surpasses the best results in \cite{zheng2021doespretraininghelpassessing}, 5 of the 7 results in \cite{chalkidis2022lexgluebenchmarkdatasetlegal} (ranging from 0.708 to 0.754) and 3 of the 5 models presented in \cite{niklaus2024_legalpile} (ranging from 0.717 to 0.770). These models do this without ``fine-tuning'' or domain adaptation. This means that general purpose LLMs may now take on tasks that historically required expensive and sophisticated fine-tuning.  

    \item As shown in Table \ref{tab: matchrate}, model accuracy degrades across all models tested for questions where the LLM answers diverge. This suggests that these questions may be more ambiguous or more complex - in future work, we will experiment with ``mixture of expert'' techniques to synthesize these varying answers into a single best answer.
\end{itemize}

\section{Have LLMs memorized the test set?}
\label{sec:memorization}

Modern LLMs are trained on an incredibly broad corpus of text, including judicial opinions. As \cite{zheng2021doespretraininghelpassessing} note, there is a risk that strong performance on these kinds of tasks may stem from ``attending to only a key word (e.g. case name)'', and suggested future research to ``disentangle memorization of case names''.  We share these concerns, and aim to ensure our results do not simply reflect rote memorization of details like the case name. This is a difficult question to answer with complete certainty, but we propose a novel experimental scheme to detect rote memorization of the case details. 

To do so, we employ a two-step prompting scheme. In the first step, we use Prompt \ref{fig:prompt2} with the \texttt{GPT4o-mini} model to anonymize the citation in question, replacing case names and citations with similar but artificial citations. The resulting text is semantically similar to the original, but modifications ensure that the text is novel and that the citations are meaningless. An example of such an original and modified citation prompt is presented in Figure \ref{fig:anonymized}. Note that all names and citations have been replaced with plausible alternatives, while the underlying facts and structure is unchanged. 

\begin{figure}[H]
\begin{shaded}
\raggedright
\footnotesize 
Input: \texttt{citing prompt}\\
    Task: Please rewrite the input text while:\\
    1. Replace all case names CONSISTENTLY:\\
       - Use different but plausible names\\
       - If a name appears multiple times, use the same replacement each time\\
       - Example: If "Smith" becomes "Wilson", all instances of "Smith" should become "Wilson"\\
    
    2. For each citation:\\
       - Change all numbers (years, page numbers, etc.)\\
       - Change the jurisdiction (e.g., F.3d → P.2d or N.Y.S.2d → Cal.App.)\\
       - Keep citations in a valid legal format\\
    
    3. Preserve exactly:\\
       - The <HOLDING> tag location\\
       - All punctuation\\
       - All legal reasoning and discussions\\
       - The basic sentence structure\\
    
    Change ALL identifying information consistently while keeping the legal meaning identical. Surround your response with tags, <output> text </output>.\\
    Output: 
\end{shaded}
\captionsetup{name=Prompt}
\caption{This prompt asks an LLM to read the citing text and replace all citations with similar but artificial values.}
\label{fig:prompt2}
\end{figure}

This procedure introduces a substantial amount of change in the citing prompts themselves: the median anonymized prompt has a Levenshtein distance of 91 edits from the original, changing roughly 10\% of the prompt. Other scholars have shown that even very small changes in the prompt, such as the inclusion of an additional space character, can induce large changes in LLM responses, so we believe that these modifications are enough to substantially alter the text from anything seen in model training\cite{salinas2024butterflyeffectalteringprompts}. Furthermore, this procedure ensures that case names are modified, removing an obvious mechanism for memorization. All original and anonymized citation texts are available for inspection at \href{https://github.com/chuck-arvin/CaseHOLD2025}{our Git repository}.

We pass this modified citing text through Prompt \ref{fig:prompt1}. Because we have not changed any core facts or reasoning, LLMs should reach the same conclusion about the text and choose the same completion option as before. However, if it turns out that the LLM is simply relying on memorization, the LLM may produce different conclusions as it can no longer rely on the memorized answer.

Despite introducing large changes to the inputs and generating fictitious legal precedents, this procedure introduces little change in the quality of the LLM outputs. The macro F1 score remains strong, going from 0.744 in the original data to 0.728 in the new data. Answers are unchanged in 88\% of cases. Though this is not conclusive, this experiment gives us some confidence that the strong results on this task are not due to memorization of the case law. 
\begin{figure}
\begin{mdframed}
\begin{minipage}{0.45\textwidth}
\footnotesize defects,” specifically, defects in establishing citizenship for the purpose of establishing diversity jurisdiction. Id. at 223. See also Harmon v. OKI Sys., 115 F.3d 477, 479 (7th Cir.1997) (citing with approval the reasoning in In re Allstate that “a defendant’s failure to allege citizenship as opposed to residency ... constituted a procedural defect”). We agree with the Fifth Circuit’s interpretation of § 1447(c) and construction of a party’s failure to establish citizenship in its notice of removal as a procedural defect. “[Wjhere subject matter jurisdiction exists and any procedural shortcomings may be cured by resort to § 1653, we can surmise no valid reason for the court to decline the exercise of jurisdiction.” In re Allstate, 8 F.3d at 223. See also Ellenburg, 519 F.3d at 198 (<HOLDING>). Section 1653 provides that “[d]efec-tive
\end{minipage}\hfill
\begin{minipage}{0.45\textwidth}
\footnotesize defects,” specifically, defects in establishing citizenship for the purpose of establishing diversity jurisdiction. Id. at \textcolor{red}{456}. See also \textcolor{red}{Taylor v. XYZ Corp., 123 P.2d 789, 791 (9th Cir.2001}) (citing with approval the reasoning in In re \textcolor{red}{Nationwide} that “a defendant’s failure to allege citizenship as opposed to residency ... constituted a procedural defect”). We agree with the \textcolor{red}{Sixth} Circuit’s interpretation of § 1447(c) and construction of a party’s failure to establish citizenship in its notice of removal as a procedural defect. “[W]here subject matter jurisdiction exists and any procedural shortcomings may be cured by resort to § 1653, we can surmise no valid reason for the court to decline the exercise of jurisdiction.” In re \textcolor{red}{Nationwide, 9 P.3d at 456}. See also \textcolor{red}{Johnson, 620 P.2d at 345} (<HOLDING>). Section 1653 provides that “[d]efec-tive
\end{minipage}
\caption{Original and ``anonymized'' citation prompts with changes in \textcolor{red}{red} (Levenshtein distance = 91).}
\label{fig:anonymized}
\end{mdframed}
\end{figure}
\section{Conclusion}

In this paper, we have explored how well modern LLMs are able to identify the correct case holding for a given legal prompt. Our results demonstrate promise for the application of LLMs in legal NLP tasks. We show that performance scales with model size, and that general purpose LLMs can now match or surpass the performance of custom-built legal models without requiring domain-specific training. Through our novel citation anonymization procedure, we demonstrate that these strong results remain after we introduce substantial changes in the input prompts, suggesting the models are doing more than mere memorization of case names. 

This work opens up a variety of future research directions. First, given the relationship between model size and performance, we expect that continued advancement in frontier-class LLMs may achieve stronger performance on these legal NLP tasks. We hope to test this effect across a wider variety of legal NLP datasets. We also plan to continue researching techniques to improve our ability to detect LLM memorization. While we believe our citation anonymization procedure is a reasonable diagnostic of LLM memorization, further improvements to our approach may yield more useful tools for detecting when LLMs are relying on memorization.

\bibliographystyle{ACM-Reference-Format}
\bibliography{references}
\end{document}